\documentclass[final]{cvpr}

\usepackage{times}
\usepackage{epsfig}
\usepackage{graphicx}
\usepackage{amsmath}
\usepackage{amssymb}
\usepackage{xcolor}

\usepackage[pagebackref=true,breaklinks=true,colorlinks,bookmarks=false]{hyperref}

\def\cvprPaperID{14} 
\def\confYear{CVPR 2021}

\begin{document}

\title{Spatio-Temporal Context for Action Detection}
\author{Manuel Sarmiento Calderó, David Varas, Elisenda Bou-Balust\\
 \\
Apple}

\maketitle

\begin{abstract}
   Research in action detection has grown in the recent years, as it plays a key role in video understanding. Modelling the interactions (either spatial or temporal) between actors and their context has proven to be essential for this task. While recent works use spatial features with aggregated temporal information, this work proposes to use non-aggregated temporal information. This is done by adding an attention based method that leverages spatio-temporal interactions between elements in the scene along the clip. 
   
   The main contribution of this work is the introduction of two cross attention blocks to effectively model the spatial relations and capture short range temporal interactions. 
  
   Experiments on the AVA dataset show the advantages of the proposed approach that models spatio-temporal relations between relevant elements in the scene, outperforming other methods that model actor interactions with their context by +0.31 mAP.
\end{abstract}

\section{Introduction}

Action detection is a task that consists in detecting people and recognizing their actions along videos. Being fundamental to video understanding, action detection has gained attention in recent years \cite{ava}, \cite{slowfast}, \cite{acar}, leading to remarkable advances.

Recent efforts in this field \cite{acar}, \cite{contextaware}, \cite{aia} are focused on modelling actor relations together with their context in the scene. These approaches have proven the advantage of incorporating relation modelling in action recognition systems, specially for understanding actions that involve actor interactions. These interactions may consider the context \cite{contextaware}, other people \cite{acar}, long term temporal relations \cite{longterm}, scene objects \cite{aia} and even audio \cite{audioslowfast}. 

\begin{figure}[t]
\begin{center}
    \includegraphics[width=1\linewidth]{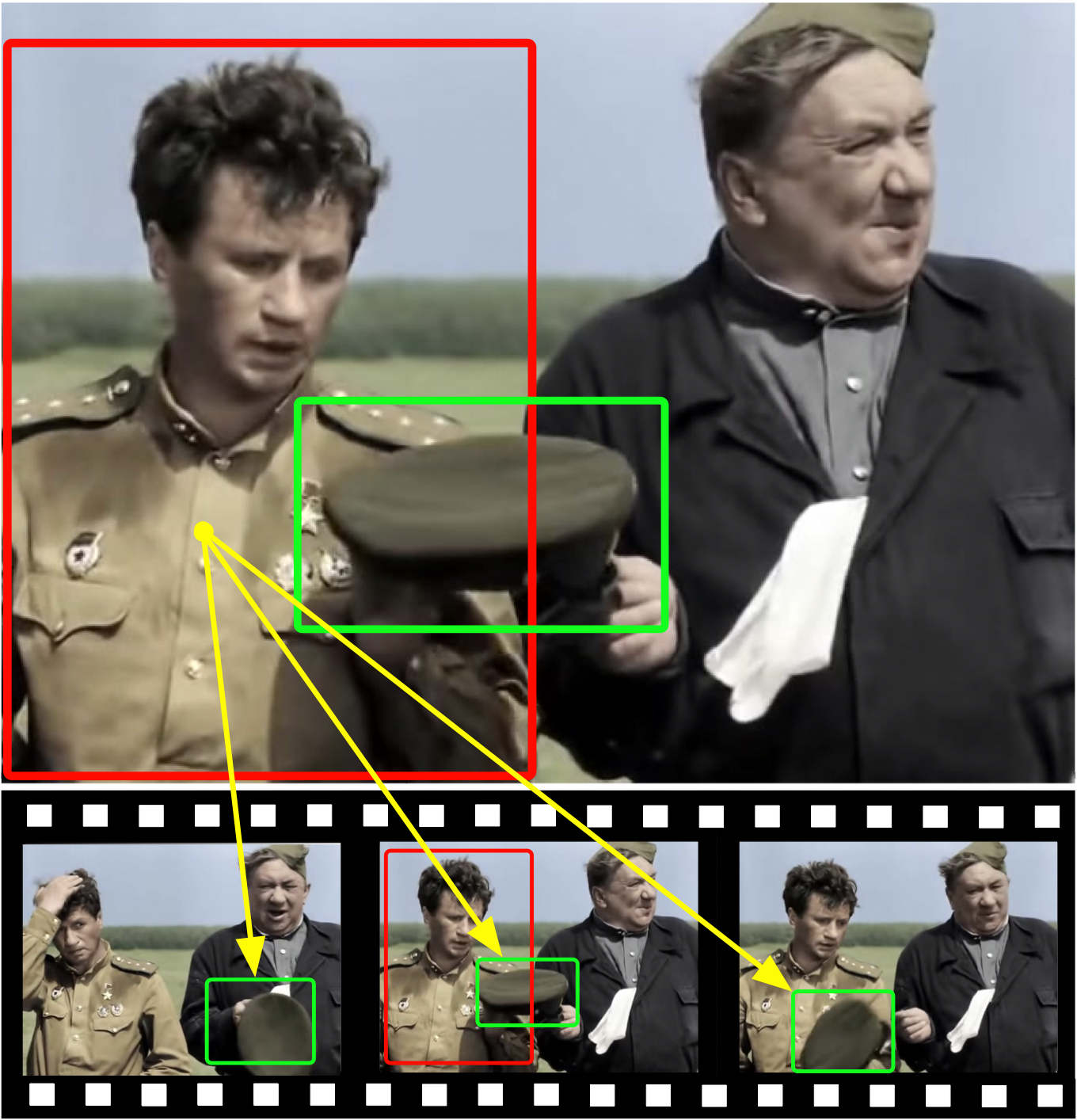}
\end{center}
   \caption{Reasoning on actor interactions only using spatial information may not be enough for understanding actions. The proposed method incorporates spatial and temporal information to robustly identify relevant features to recognize the action performed by the actor (red) with the hat (green).}
\label{fig:context_example}
\end{figure}

In order to model these interactions, it is important to take into account both spatial and temporal information. While long term temporal information is currently leveraged using a memory system over the entire video clip\cite{longterm}, this work proposes to add temporal information present in adjacent frames directly in the relation modelling of actor interactions. As shown in figure \ref{fig:context_example}, it is hard to infer the correct interaction between the actors appearing a video sequence only considering the central frame. Specifically, using only this information it is a difficult task to know whether the young man in the red bounding box is giving or receiving the hat from the old man. In this kind of actions, adjacent frames contain crucial information to explain what is happening in the video. However, previous approaches aggregate temporal information before modelling interactions between actors and other elements of the scene \cite{acar}, \cite{contextaware}. This may result in losing part of this temporal information, which is key for understanding relations.

This paper leverages action features using a combination of spatial and temporal information to model interactions between the actor and the context for action detection. 
To this end, a new method that uses an attention mechanism and treats spatial and temporal information separately is proposed. The resulting system is able to exploit the most relevant parts of each type of feature. 
Therefore, the main contribution of this paper is the introduction of a novel attention-based architecture that uses factorized spatial and temporal information to understand these interactions. 

This paper is organized as follows: Section \ref{sec:relatedwork} reviews previous works in the field of action recognition that model contextual information. Then, in Section \ref{sec:method} the proposed model is presented. Section \ref{sec:experiments} follows with the experimental details, dataset and hyperparameters for all the experiments. Finally, in Section \ref{sec:conclusions} the conclusions of this work are drawn.

\section{Related work}
\label{sec:relatedwork}
In this section, previous methods focused on action detection using contextual information are reviewed.

Early works in the action recognition field were focused on classifying short trimmed videos \cite{kinetics} using complex 2D architectures. The appearance of 3D CNNs \cite{c3d, slowfast} drastically improved the performance of previous methods in many action recognition tasks such as action detection \cite{betterava}.

Action detection, which is the task of assigning an action to each actor in a video, has also been impulsed by the appearance of datasets with densely annotated atomic actions \cite{ava}.
In \cite{betterava, slowfast}, a common pipeline for action detection with a backbone pretrained using \cite{kinetics} and a person detector is presented. However, these approaches do not consider any reasoning on actor interactions.


One of the first works that takes into account interactions with parts of the scene is \cite{nonlocal}. This work, learns to capture long range interactions between the extracted features. 
Similarly, \cite{videoactiontransformernetwork} tries to model contextual information with a transformer architecture \cite{transformer} using the actor RoIs as attention queries. Recently, other type of interactions such as long term temporal relations \cite{longterm} or relations between audio and video \cite{audioslowfast} have also been considered. 

While relations between an actor and his context are considered first-order relations, other works explore adding the relationship between first-order relations as a second-order relation (e.g. actor-context-actor).
In \cite{acar}, these second order relations between actors are incorporated on top of context interactions. Following the idea of using more than one type of relation, \cite{aia} leverages action features with long term temporal and object features.

The work presented in this paper aims at capturing relations between actors and the scene, differing from previous works in the use of non-aggregated spatial and temporal information for modelling actor interactions. While this work focuses on adding first-order relations, second-order could also be added in the future.

\section{Method}
\label{sec:method}

In this section, the proposed method to combine context features for action detection is introduced. 
The architecture uses an attention mechanism to enrich actor features with spatio-temporal information extracted from the scene. First, an overview of this system is presented. Then, the use of spatio-temporal features for context modelling using a cross attention architecture is described in detail. 

\subsection{Overview}
\label{subsec:overview}
Following common action detection frameworks \cite{slowfast, longterm}, the proposed system is designed to detect people in videos and estimate one or more actions for each of them. To this end, an attention based method is used to perform relation modelling between actors and scene elements of the clip. 

\begin{figure}[t]
\begin{center}
    \includegraphics[width=1\linewidth]{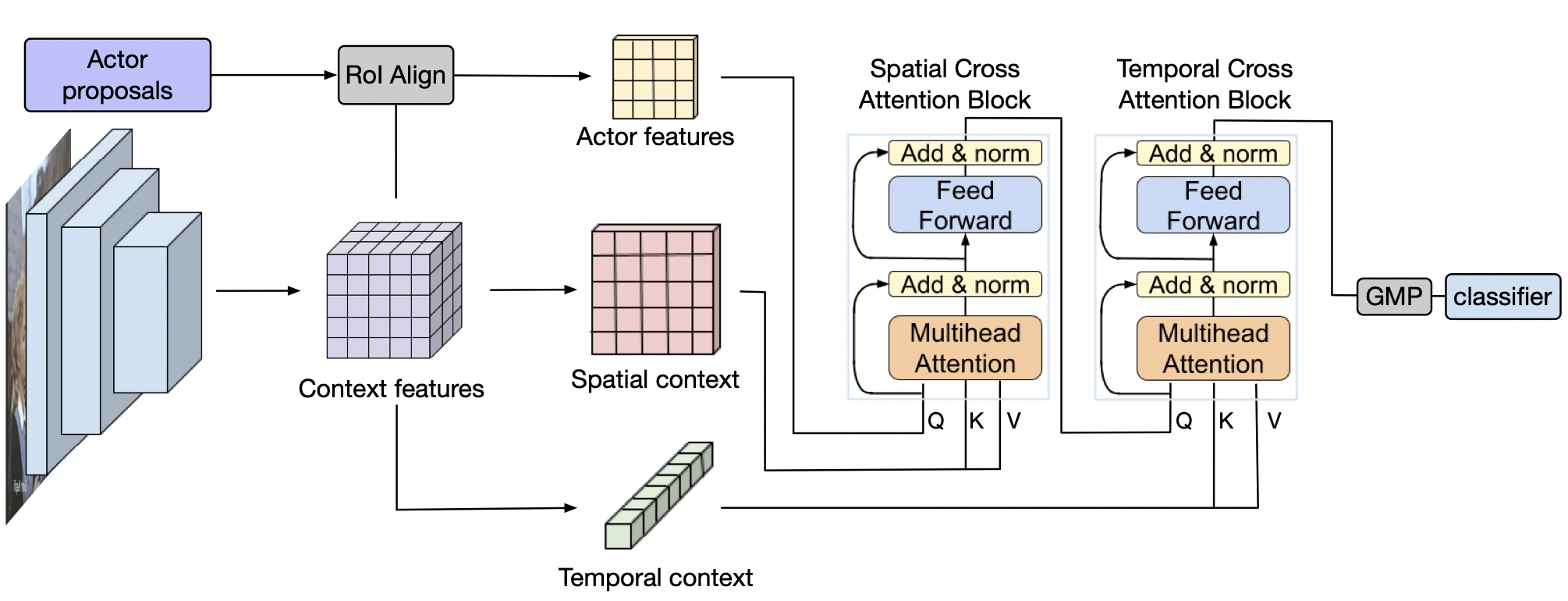}
\end{center}
   \caption{System overview. The system presented in this paper employs two cross attention blocks to enrich the actor features with spatial and temporal relations from the clip. The output feature is finally used for action classification.}
\label{fig:system_overview}
\end{figure}

First, actor proposals from a person detector are combined with a backbone feature extractor to obtain actor features (Figure \ref{fig:system_overview}). The person detector is only used on the central frame of the clip to obtain $N$ actor bounding boxes. At the same time, the video clip is passed through the backbone \cite{slowfast}, that extracts two feature maps $(F_s, F_f)$ of shape  $T_s\times H\times W\times C_s$ and $T_f\times H\times W\times C_f$. To obtain the actor features, an average temporal pooling is performed to both feature maps. Then, these pooled features are concatenated and a RoI Align operation is applied to extract $N$ features $(A_1, ...,  A_N)$ of shape $7\times 7\times C$ each.

Then, a cross attention block, which consists of a multihead attention and a feed forward layer, \cite{transformer} enriches these $N$ actor features with the spatio-temporal information of the backbone features $(F_s, F_f)$ to model interactions between the actors and their context. The resulting features are spatially reduced with a global maximum pooling (GMP) and passed through two linear layers to predict the desired output.

\subsection{Context Modelling}
\label{subsec:context_modelling}
The context in which an action takes place may be crucial for its understanding as actors usually interact with parts of the scene while performing this action. 

In order to capture the relations between actors and the context in the scene, the proposed system incorporates a cross attention mechanism based in the Transformer. The cross attention queries (Q) are obtained from the actor features, similar to \cite{videoactiontransformernetwork}. On the other hand, backbone features are used as keys (K) and values (V). In practice, for each actor, the cross attention mechanism finds the relevant context information in the scene.

\textbf{Actor Features.} 
Previous works propose to reduce the spatial dimension of actor features before modelling interactions with the context \cite{videoactiontransformernetwork, acar, contextaware}. However, these spatial dimensions may be key to understand how an actor is interacting with its surroundings (e.g. the hand of the actor taking an object). For this reason, in this work spatial pooling is not performed to the actor features to effectively model the spatial relations between actors and scene features.



\textbf{Spatial and temporal cross attention blocks.} 
The backbone used in this work is composed of two pathways \cite{slowfast}. On the one hand, a slow pathway that operates at low frame rate to capture spatial semantics of the scene. On the other, a fast pathway with a high frame rate captures motion at fine-grained temporal resolution. Slow features $F_s$ can be interpreted as features that gather the spatial information while fast features $F_f$ can be interpreted as the temporal information of the sequence.

Due to the different nature of these features, a system that combines both slow and fast features extracting the relevant contextual information from each one is designed. This system uses both types of features to provide specific context information to the attention mechanism. 

The proposed system uses the contextual information dividing the spatial and temporal dimensions (Figure \ref{fig:system_overview}). This is done using two attention blocks. First, actor features are enriched with the spatial information of slow features. As $F_s$ contain mainly spatial information, a temporal pooling is performed, preserving the spatial dimensions. Then, actor features are used as queries and temporal pooled slow features as keys and values of the first cross attention block.

Second, the relevant information of the fast features is incorporated to these enriched features. In contrast to the temporal pooling of $F_s$, a spatial pooling is performed to $F_f$ to preserve the temporal information. The previously enriched actor features are used as keys while the spatial pooled fast features are used as keys and values of the second cross attention block. Using this mechanism, the attention can select the relevant information of the fast features that has not been previously extracted from the slow features. Thus, specific information of each type of feature can be used to enrich actor features.

\section{Experiments}
\label{sec:experiments}

In this section, the proposed architecture is assessed using the version 2.2 of the AVA dataset \cite{ava}. AVA is a video dataset for spatio-temporal localization of atomic human actions. This dataset is composed of 430 video clips of 15 minutes extracted from movies and the actor annotations are provided at one frame per second. Each annotation consists of a bounding box and multiple action labels. All the results provided in this work are obtained using the 60 most common classes of this dataset, following the standard evaluating protocol \cite{ava}. The performance metric is mean Average Precision (mAP) using a frame-level IoU threshold of 0.5.

\subsection{Implementations details}

\textbf{Person Detector.} Actor detections used in this work are precomputed bounding boxes from \cite{slowfast}. Following its procedure, both the ground truth boxes and the detected actor boxes with confidence higher than 0.8 are used as proposals.

\textbf{Backbone Network.} The backbone is a SlowFast R50 with input sampling $T \times \tau = 8 \times 8$, without non local blocks and spatial resolution of res$_5$ increased by 2. The input are 32 frames with temporal stride of 2 for the fast pathway, and 8 frames with temporal stride of 8 for the slow pathway.


\textbf{Training and Inference.} For training the backbone, the standard procedure of initialising the backbone with the weights from the Kinetics-400 \cite{kinetics} classification model is used. During training, the network inputs are $T \times \tau$ frame crops of size 224$\times$224, and during inference, $T \times \tau$ frames of short size 256 pixels are employed.

\subsection{Ablation study}

In this section, the importance of the spatio-temporal features for relation modelling is analyzed. A backbone SlowFast R50 8x8 with an actor detector and a classification layer as in \cite{slowfast} is considered as baseline. All the experiments in this section use the same backbone. 

Two main aspects of the system are analyzed. First, the context features used for obtaining actor interactions. For this assessment, using only temporal pooled features (\textit{Spatial Context}) or a combination of temporal and spatial pooled features (\textit{Spatial+Temporal Context}) is considered. The latter is the approach presented in Section \ref{sec:method}. Second, the actor features dimensionality. In addition to the common pooled features used in previous works, the use of these features keeping their spatial dimensions (\textit{Spatial)} is analyzed. 

The network of the experiments using \textit{Spatial Context} has a single cross attention block which keys and values consist in a concatenation of SlowFast backbone feature maps ($F_s$, $F_f$) after a temporal pooling. 
In the \textit{Spatial+Temporal Context} experiment, a temporal pooling over the slow features is used for the spatial cross attention and a spatial pooling on the fast features is used for the temporal cross attention block as described in Section \ref{subsec:context_modelling}.

The results of these experiments are presented in Table \ref{tab:spatiotemporal}. As it can be observed, incorporating temporal information to the context features improves over using only spatial information. Moreover, using actor features with spatial dimension also results in an better performance in both the \textit{Spatial Context} and the \textit{Spatial+Temporal Context} experiments. 
This proves that the proposed system extracts relevant information from the interactions of spatio-temporal features to understand actor relations.
 

\begin{table}[h]
    \centering
    \begin{tabular}{|l|c|c|}
    \hline
    Method & Actors Features & mAP \\
    \hline\hline
    Baseline & - & 24.80 \\
    Spatial Context & - & 26.50 \\
    Spatial Context & Spatial &  26.75 \\ 
    Spatial+Temporal Context & - &  26.65 \\
    Spatial+Temporal Context & Spatial &  27.02 \\ 
    \hline
    \end{tabular}
    \caption{Effect of spatiotemporal features for relation modelling.}
    \label{tab:spatiotemporal}
\end{table}

\subsection{Comparison with state of the art}

Results obtained by the proposed system compared with state of the art methods are set in Table \ref{tab:stateoftheart}. In this table, only methods pretrained on Kinetics 400, that use similar backbones are compared. The comparison is made with other works that take into account relations with the scene. Moreover, systems that model long term temporal relations or use audio features are also considered for this comparison. In the case of ACAR-Net, the comparison is made against the system that models first order interactions because the introduced system in this paper does not take into account actor-actor interactions. Nonetheless, note that any system that models other type of interactions can be used on top of the proposed system. 
These results show that the proposed system is able to extract relevant information from the non-aggregated spatial and temporal features an achieve a 27.02 mAP, outperforming the previous methods. 


\begin{table}[h]
    \centering
    \begin{tabular}{|l|c|c|c|}
    \hline
    Model & Backbone & AVA & mAP \\
    \hline\hline
    SlowFast \cite{slowfast} & R50 8$\times$8 & 2.1 & 24.80 \\
    Action Tx \cite{videoactiontransformernetwork} & I3D & 2.1 &  25.00 \\
    LFB \cite{longterm}& R50 & 2.1 & 25.80 \\
    Context-Aware RCNN \cite{contextaware}& R50 16$\times$4 & 2.1 & 25.80 \\
    AVSlowFast \cite{audioslowfast} & R50 4$\times$16 & 2.2 & 25.90 \\
    ACAR-Net \cite{acar} & R50 8$\times$8 & 2.2 & 26.71 \\ 
    Ours & R50 8$\times$8 & 2.2 &  27.02 \\
    \hline
    \end{tabular}
    \caption{Comparison with state of the art on AVA v2.2 validation set.}
    \label{tab:stateoftheart}
\end{table}

\section{Conclusions}
\label{sec:conclusions}

This work presented a novel system that leverages temporal information from adjacent frames together with spatial information to improve the recognition of actor interactions in video clips.

The proposed method combines spatial actor features with spatio-temporal context feature maps to output an enriched feature for action classification. This enhancement is performed with an architecture that uses two cross attention mechanisms to extract the most relevant parts of spatial and temporal information.

As shown in the experiments section, this system outperforms state of the art methods that use scene information to contextualize actor features. The assessment was performed in the AVA v2.2 dataset resulting in an improvement of 0.31 points in mAP@0.5. These results open the door towards the usage of temporal features in contextualized action detection.

{\small
\bibliographystyle{ieee}
\bibliography{egbib}
}

\end{document}